**Decolonising Data Systems: Using Jyutping or Pinyin as tonal representations of Chinese names for data linkage**

Joseph Lam*[1], Mario Cortina-Borja[1], Robert Aldridge[2], Ruth Blackburn[1], Katie Harron[1]

Affiliations

*Corresponding Author

1. Great Ormond Street Institute of Child Health, University College London, UK

2. Institute for Health Metrics and Evaluation, University of Washington, USA

Word Count: 2,987

**Abstract**

Data linkage is increasingly used in health research and policy making and is relied on for understanding health inequalities. However, linked data is only as useful as the underlying data quality, and differential linkage rates may induce selection bias in the linked data. A mechanism that selectively compromises data quality is name romanisation. Converting text of a different writing system into Latin based writing, or romanisation, has long been the standard process of representing names in character-based writing systems such as Chinese, Vietnamese; and other languages such as Swahili. Unstandardised romanisation of Chinese characters, due in part to problems of preserving the correct name orders the lack of proper phonetic representation of a tonal language, has resulted in poor linkage rates for Chinese immigrants. This opinion piece aims to suggests that the use of standardised romanisation systems for Cantonese (Jyutping) or Mandarin (Pinyin) Chinese, which incorporate tonal information, could improve linkage rates and accuracy for individuals with Chinese names. We used 771 Chinese and English names scraped from openly available sources, and compared the utility of Jyutping, Pinyin and the Hong Kong Government Romanisation system (HKG-romanisation) for representing Chinese names. We demonstrate that both Jyutping and Pinyin result in fewer errors compared with the HKG-romanisation system. We suggest that collecting and preserving people's names in their original writing systems is ethically and socially pertinent. This may inform development of language-specific pre-processing and linkage paradigms that result in more inclusive research data which better represents the targeted populations.

We hope this piece will increase awareness of the extent of data linkage problems that differentially impact communities globally and provide suggestions to work towards active decolonisation of data, by inspiring alternative ways of valuing, capturing and utilising information in data systems.

**Introduction: Differential Linkage Errors Through Romanisation**

Data linkage is used increasingly in large administrative data systems to enhance data resources to answer more complex research and policy questions (1). Accurate data linkage is important as linkage errors can undermine the quality of the linked data by introducing information and selection bias, limiting researchers' ability to accurately represent the targeted populations (2). In data linkage, names are often used as key identifiers to decide if records from multiple data sources belong to the same person.

Romanisation refers to the transformation of names in local languages to commonly operable language, which is English in most developed countries in the Global North. Within data systems, romanisation can be viewed as a process of data reduction that selectively retains information operable in (Western) data systems (e.g., only containing Latin-based alphabets), and drops information deemed irrelevant or too costly to retain. For example, tones in tonal languages are often not retained, people with multiple surnames (e.g., common in Spain and Latin American countries) might only be allowed to provide one or use hyphenated versions, and letters not in the English alphabet and diacritical marks might be ignored or misrepresented. Romanisation is not always standardised, e.g., there are over 10 different systems each for Cantonese, Korean, and Arabic.

When romanised names are not consistently represented in data systems, linkage errors are more likely to occur (i.e. false matches, where records belonging to different people are linked together, or missed matches, where records belonging to the same person remain unlinked). As large-scale linked administrative data are used more readily by health professionals and policy makers to generate evidence and drive decisions, this disparity in linkage rates perpetuates health and social inequities for under-represented groups, such as migrants and people from ethnic minoritised communities (3). For example, a historical linkage of census data from the United States matched only 3.6% of male Chinese migrants between 1880-1900 compared to 16.3% of English migrants (4). A recent linkage of asylum and resettled refugees with census data in the United Kingdom found a substantial difference in linkage rates by language and country of origin (5). There is a growing proportion of births in England (37%) and London (66%) to families where one or both parents were born outside the UK (6); between 2013-2017, 53% of singleton births in New York City were born to non-US-born mothers (7). Use of poorly romanised and processed names in data systems will selectively and continually leave parts of the population behind, if current data linkage 'blind spots' or structural barriers in data systems to inclusion in linked datasets are not actively addressed.

This opinion piece describes three problems faced by, but not unique to representing Chinese characters for data linkage. We compare standardised romanisation of Cantonese (Jyutping), Mandarin (Pinyin), with the non-standardised Hong Kong Government Cantonese Romanisation system, as a case example. We then propose a solution model for decolonising data systems, by demonstrating the advantages of challenging current value positions and priorities in global data systems to improve linkage and promoting data equity (8).

> In modern Chinese, a full name starts with a surname, followed by a forename. There are two writing systems: Traditional Chinese, used by people in Taiwan, Hong Kong and Macau; and Simplified Chinese, used mainly by people in China and by people of Chinese heritage in other South Asian countries. Traditional and Simplified Chinese differ only by writing but share the same pronunciations.
>
> Chinese is a character-based language. Surnames typically consist of one to two characters, and up to nine characters (9). The Grand Dictionary of Chinese Surnames (9) collated 11,969 Surnames, where over 90% of the Chinese population shared 120 common surnames and all of them consist of one character. The top five surnames (Wang, Li, Zhang, Liu, Chen) account for over 30% of the population (10). Forenames usually consist of one to two characters, with no upper limit. In 2020, over 90% of Chinese full names consisted of 3 characters, whilst only around 6% of full names had 2 characters, and 3% had 4 or more characters. Modern Chinese surnames are patrilineal, where infants have their father's surnames. However, about 8% of newly registered babies had their mother's surnames (10).
>
> Standard Chinese, or Mandarin, is spoken by about 80% of the population in China; Yue Chinese, or Cantonese, is the second most spoken Chinese language with over 80 million native speakers in Hong Kong, Macau, and Guangdong, Guangxi provinces. Cantonese is also predominantly spoken by ethnic Chinese communities in Vietnam, and early Chinese migrants in Europe and North America. There are other dialects of the Chinese language. This piece focuses on Cantonese and Mandarin as they represent the majority of Chinese speakers.

Box 1. Introduction to Chinese names

**Chinese Romanisation systems**

**Hong Kong Government Cantonese Romanisation (HKG-romanisation) - Cantonese**

Hong Kong has used romanised Cantonese to represent places and official names, and for documentation, from the British colonial periods (1800s) to present (11). The romanisation system used by the Hong Kong Government was based on three legacy systems developed in the 19th century British missionaries in Hong Kong and China and it remains the official system. Linguists have since highlighted substantial inconsistencies in representing consonants, vowels, diphthongs and syllabic consonants within the HKG-romanisation system (12). For example, the vowel "ei" represented with International Phonetic Alphabet can be represented as "ei", "ee", "ay", "ai" or "i" using HKG-romanisation. The same problem manifests in names.

For example, a frequent surname "楊" can be represented as "yang", "young", "yep", "yong", "yeung", "yeang", "yung". HKG-romanisation does not capture tones. Each character is separated by a blank space.

**Jyutping – Cantonese**

Jyutping is a romanisation system developed in 1993 by the Linguistic Society of Hong Kong (13). Jyutping represents Cantonese with 6 tones. It is capable of representing all modern Cantonese sounds and tones with only alphanumeric characters without any diacritics or other symbols: the same example surname "楊" will only be represented as "joeng4". Each character is separated by a blank space.

**Pinyin - Mandarin**

Pinyin was developed in 1950s by a group of Chinese linguists as part of a National Reform of the Chinese Written Language (14). Pinyin represents Mandarin with 4 tones plus a neutral tone. Tones are written as diacritics, ā, á, ǎ, à, and a – neutral tones are presented without any accent marks. As per the Official Basic Rules of the Chinese Phonetic Alphabet Orthography (15), surname and forename are separated by a blank space, but sur/forename with multiple characters are not separated by a blank space. For ease of comparison, Pinyin tones are represented using numbers 1,2,3,4 and 5.

**Three Problems with processing Chinese characters and names**

1) Language-specific Romanisation

Theoretically, Jyutping or Pinyin should more accurately represent pronunciations compared to the HKG-romanisation system (Box 1). However, not all variations in romanised representations arise from flaws in the romanisation system. Instead, there are historic-, country- and language-specific variations in how some Chinese characters are represented in different geographic regions and countries – thus this variation can be informative for linkage. Figure 1 Illustrates the variations in representation of a common surname "林" in Cantonese, Mandarin, Vietnamese, Malaysian/Singaporean, Indonesian, Japanese, Korean, and other dialects. These variations of "林" are not a result of inconsistent romanisation, but a reflection of how that character is pronounced locally, and how such pronunciations change over time within the same region. Using a single unified romanisation system for Chinese characters from all countries would mean these language-specific and temporal-specific distinctions might be lost. Knowing the country of origin and language system of the individuals may help identifying the best romanisation approach to retain most relevant information.

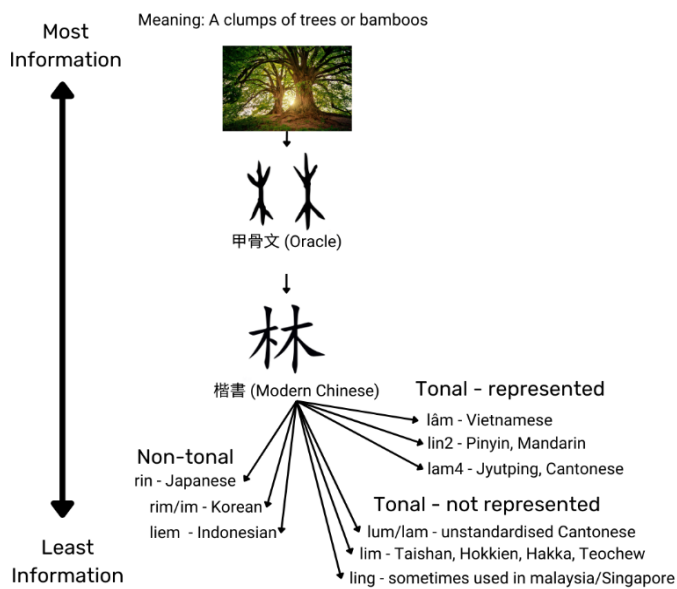

*Figure 1*. Depiction of different ways the character '林' is romanised and pronounced by countries and languages.

Photo by Johannes Plenio: https://www.pexels.com/photo/two-brown-trees-1632790/.

2) Non-tonal representation of a tonal language

Modern data systems were developed in mind for Indo-European languages using Latin-based alphabets. With rare exceptions such as Panjabi (Punjabi), Indo-European languages do not differentiate tones in how they are spoken. This means that even if local languages were perfectly represented by roman alphabets, romanised names will still be less specific and distinctive than their original names as tonal information is not retained. This loss of granularity degrades the identifying information available and could lead to linkage errors. This is a particular problem for HKG-romanisation and Pinyin, as intonations are not represented in the former, and not properly recorded in the latter as diacritics are often ignored. Jyutping and a properly recorded Pinyin system are more sensitive to tonal changes, and a more consistent representation of Chinese characters than HKG-romanisation.

3) Name orders

A common issue for character-based languages is the misplacement of characters. For example, Chinese surname and forename are often inverted due to different naming conventions; forename characters are sometimes misplaced as middle names. The latter is more prevalent in HKG-romanisation and Jyutping, as each character is separated by a space; and less so for Pinyin is as there is no space within forename characters. The inability to segment which characters belong to surname or forename fields means that linkages may be inaccurate. Previous attempts, e.g., the ABE method (4), to pre-process multi-part names have resorted to clustering multi-part names to their first character, e.g., clustering "Chin Fung", "Chin Hing", "Chin Lung" as "Chin". Note that these are all HKG-romanisation, linking Cantonese-based romanised names would be predominantly impacted using the ABE method. Chinese surnames are already less specific for linkages than English, with larger clusters of people sharing same common surnames. The ABE method further lowers the specificity of Chinese forenames, resulting in linkages that are more prone to false matches. Postel (16) demonstrated that proper segmentation, indexing and ordering of Chinese characters could substantially improve linkage rates for Chinese names.

In the following section, we will compare the utility of Jyutping, Pinyin and HKG-romanisation in representing Chinese characters, sensitive to language-specific romanisation, tonal representation and name orders.

**Experiment**

We scraped online student class lists from schools in Hong Kong that provide both Chinese and English Names (*n* = 774). We only included names that had a Mandarin or Cantonese origin, based on the provided Chinese and (romanised) English names (n = 771). Records providing English names with no space within forename characters, or use of non-accented Pinyin as English names reflect a Mandarin origin.

We cleaned, segmented and re-ordered the data, then derived Jyutping and Pinyin using the pinyin_jyutping package (17) in Python 3.8 (18), which used an online open-source Cantonese dictionary CC-Canto (19), for each character of the Chinese name. We manually entered the records that the package failed to translate, and stored pronunciations and tones in separate columns. Raw name list and cleaned data are available from a UCL Repository (20). Codes are available on GitHub (21).

We compared how closely 3 different systems of romanisation (HKG-romanisation, Jyutping, Pinyin) represented the original Chinese characters, in terms of uniqueness, which provided some information on the utility of these systems for balancing sensitivity and specificity of linkages.

Table 1. Descriptive characteristics of Chinese and English names in the study dataset

| Descriptive Characteristics | | |
|---|---|---|
| | Count | % |
| Total *n* | 771 | 100 |

| | | | |
|---|---|---|---|
| Chinese surname | | | |
| | 1 Character | 771 | 100 |
| Chinese forename | | | |
| | 1 Character | 17 | 2.3 |
| | 2 Characters | 754 | 97.7 |
| English forename | | | |
| | Romanised only (e.g., Chin Hang) | 647 | 83.9 |
| | English only (e.g., Johnny) | 27 | 3.5 |
| | Both English and Romanised (e.g., Chin Hang Johnny) | 97 | 12.6 |
| Language | | | |
| | Cantonese | 751 | 97.4 |
| | Mandarin | 20 | 2.6 |

**Findings**

Of the 771 included names, a large majority (97.7%) had a 3-character full name, most (83.9%) had a romanised English forename, and most names were given based on Cantonese (97.4%) (Table 1).

Table 2. Count of unique values at each field using Chinese characters, Jyutping, Pinyin, Pinyin_notone (without tonal information) and HKG-romanisation. In brackets, degree of loss or addition of unique values, using different romanisation methods compared to original Chinese characters (%).

| Unique Count | Chinese | Jyutping | Pinyin | Pinyin_notone | HKG-romanisation |
| --- | --- | --- | --- | --- | --- |
| Surname (1 char) | 123 | 117 (-4.9%) | 120 (-2.4%) | 108 (-12.2%) | 152 (+23.6%) |
| Forename (1-2 char) | 743 | 642 (-13.6%) | 679 (-8.6%) | 648 (-12.8%) | 687 (-7.5%) |
| Full name | 771 | 767 (-0.5%) | 769 (-0.3%) | 763 (-1.0%) | 770 (-0.1%) |

Our sample of 771 individuals all had unique Chinese full names, and shared 123 unique surnames (Table 2). The top 5 most frequent surnames accounted for over 30% of surnames. HKG-romanisation resulted in 152 unique surnames with 29 extra surnames than Chinese. Both Jyutping and pinyin reduced the number of unique representations of names. The HKG-romanisation system represented different Chinese characters using the same codes, for example, "Chiu" is used to represent "趙" (Jyutping: Ziu6, Pinyin: Zhao4) and "邱" (Jyutping: Jau1, Pinyin: Qiu1). The HKG-romanisation also represented the same Chinese characters using different codes, for example, "周" is represented as "Chow", "Chau", or "Chiau", where Jyutping would consistently represent it as "Zau1", and Pinyin "Zhou1". Both Jyutping and Pinyin represented the same characters consistently, but both represented different characters using the same codes, for example, "Wong4" for both "黃" and "王" in Jyutping, and "Yan2" for both "颜" and "严" in Pinyin. Pinyin without tones had all the problems of Pinyin, plus the inability to differentiate tones, hence was less specific.

There were 743 unique combinations of forenames in Chinese, corresponding to 642 unique Jyutping representations, and 679 Pinyin representations (Table 2). In our data of predominantly

Cantonese-based names, we observed more unique Pinyin forename combinations than Jyutping. Similar to naming clusters in other languages, Hong Kong people tend to use variations of similarly sounding names. For example, forename "Paak3Hei1" occurred 6 times, but each represented a unique Chinese forename. Pinyin would represent these 6 names in 4 different ways. In this case, Pinyin becomes more specific than Jyutping in differentiating people in our sample. We expect the specificity of Pinyin to be lower in a majority mandarin-speaking sample, or in a larger population with more variations in names. Both Jyutping and Pinyin provide more utility for linkage in dealing with pronunciation-based errors and rare names.

**Informing Missingness**

A further promise of tonal representations of Chinese names is the potential of using tones to impute missing names. Frequencies of tonal combinations of Jyutping and Pinyin names likely follow a Zipf's distribution (22), where a few combinations represent most names and a long tail of combinations have very few counts. In our sample, the top 10 tonal combinations represented 38.3% names in Jyutping and 45.0% in Pinyin (Figure 2). By further incorporating vowel information, statistically estimating missing tonal information should help to calibrate non-tonal romanisation systems. For example, in a 3-character name where the first 2 characters have the Pinyin tone "2-3" and the last character is missing, we can expect it is at least twice as likely for the third character to have a "2" tone than a "4" tone. Further work using a large-scale Chinese names database would contribute significantly to this regard.

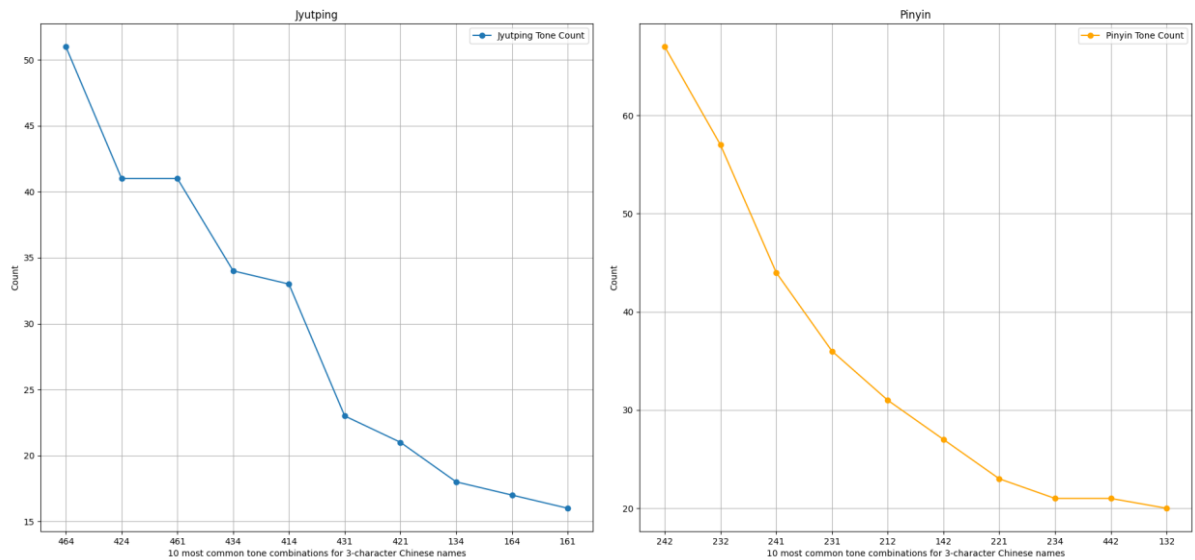

*Figure 2*. Top 10 most common tonal combination for 3-character Chinese names, in Jyutping (left) and Pinyin (right). Number on x-axis corresponds to tonal combinations.

**Implication and Conclusion**

Decolonising means to challenge the existing and often implicit value system of defining what is worth measuring, to shape it with a locally led, contextually informed and tailored approach that re-assigns values in what is captured, and how they are captured in a data system (8). Our experiment demonstrates that both Jyutping and Pinyin are promising alternatives to representing Chinese characters, accounting for pronunciation-based errors, rare names and tonal information, compared to HKG-romanisation or non-tonal Pinyin.

The recommendation when designing a linkage strategy is to always consider the data sources one intends to link, by inspecting or using expert knowledge of the particular people groups one may expect to observe in the population. For example, in the 1880-1990 US linkage, Chinese names in the US Census are more likely to be based on Cantonese than Mandarin. Blocking strategies based on Jyutping surnames may therefore increase efficiency of linkage by identifying more true matches, but will significantly increase the number of missed matches, compared to Pinyin or other romanisations. As romanised Chinese names are ascribed by immigration officers in this particular setting, Jyutping will represent forenames more precisely and sensitively than Pinyin for this linkage.

Developing a language-specific pre-processing and romanisation paradigm requires leadership and skills from people who speak these languages. We ask developing countries or former colonial regions (such as Hong Kong), where English remains the dominant language in which these data systems are operating, to consider how names can be best represented and how tonal information can be retained in data systems. As for developed countries, tone-sensitive romanisation systems should provide more flexibility in developing linkage strategies and could improve linkage quality for minoritised ethnic populations and migrants. Operationally, with vast advancement in voice-to-text transcription, asking individuals to say their name in their mother tongue may be a simple way to record extra information that is conducive to data linkage. This is relevant for other tonal languages, as well as character-based non-tonal languages.

Proper representation of people's names is integral to building trust and service delivery. In a recent Hong Kong migrant's community event hosted by the lead author, a participant complained that the National Health Service referred to them only by their surname and the first character of their forename, which is shared within their extended family. They dared not to open these letters in fear of infringing their relative's health privacy. In higher education, some educators will only refer to students by the name provided by the student registration system, which again clusters multi-term forenames to the first term.

Collecting, preserving and utilising people's names in their original language systems, or alternatively standardised romanisation systems, is ethically and socially pertinent and may support the development of language-specific pre-processing and linkage strategies that result in more inclusive research data which better represents the targeted populations.

**Ethics approval**

No ethics approval is required as only openly accessible data are used.

**Data Availability**

Raw name list and cleaned data are available on UCL Repository (20). Codes are available on JL's GitHub (21).

**Author contributions**

JL conceptualized the project, designed, analysed and wrote up the first draft. All authors contributed to critical reviewing and revising the manuscript. All authors read and approved the final manuscript before submission and agreed with the decision to submit the manuscript.

**Funding**

This work was supported by the Wellcome Trust [212953/Z/18/Z].

**Conflict of Interest**

None declared.